\pdfoutput=1
\documentclass[11pt]{article}

\PassOptionsToPackage{bookmarksnumbered}{hyperref}

\usepackage[final]{acl/acl}

\usepackage{times}
\usepackage{latexsym}
\usepackage[T1]{fontenc}
\usepackage[utf8]{inputenc}
\usepackage{microtype}
\usepackage{inconsolata}
\usepackage{graphicx}
\usepackage{amsmath}
\usepackage{mathtools}
\usepackage{amsthm}
\usepackage{cleveref}
\usepackage{color}

\title{The Impossibility of Fair LLMs}

\author{
  \textbf{Jacy Reese Anthis}\textsuperscript{1,2},\ 
  \textbf{Kristian Lum}\textsuperscript{1},\ 
  \textbf{Michael Ekstrand}\textsuperscript{3},\\
  \textbf{Avi Feller}\textsuperscript{4},\ 
  \textbf{Chenhao Tan}\textsuperscript{1}\\
  \\
  \textsuperscript{1}University of Chicago\quad
  \textsuperscript{2}Stanford University\\
  \textsuperscript{3}Drexel University\quad
  \textsuperscript{4}University of California, Berkeley
}

\begin{document}
\maketitle
\begin{abstract}
    The rise of general-purpose artificial intelligence (AI) systems, particularly large language models (LLMs), has raised pressing moral questions about how to reduce bias and ensure fairness at scale. Researchers have documented a sort of “bias” in the significant correlations between demographics (e.g., race, gender) in LLM prompts and responses, but it remains unclear how LLM fairness could be evaluated with more rigorous definitions, such as group fairness or fair representations. We analyze a variety of technical fairness frameworks and find inherent challenges in each that make the development of a fair LLM intractable. We show that each framework either does not logically extend to the general-purpose AI context or is infeasible in practice, primarily due to the large amounts of unstructured training data and the many potential combinations of human populations, use cases, and sensitive attributes. These inherent challenges would persist for general-purpose AI, including LLMs, even if empirical challenges, such as limited participatory input and limited measurement methods, were overcome. Nonetheless, fairness will remain an important type of model evaluation, and there are still promising research directions, particularly the development of standards for the responsibility of LLM developers, context-specific evaluations, and methods of iterative, participatory, and AI-assisted evaluation that could scale fairness across the diverse contexts of modern human-AI interaction.
\end{abstract}

\section{Introduction}

In response to the rapid adoption of machine learning systems and concerns about their negative societal impacts, researchers have developed compelling, nuanced technical frameworks to formalize ethical and social ideals—particularly the foundational notion of ``fairness''—in order to systematically evaluate and apply them. Popular fairness frameworks include group fairness \cite{dwork11} and fair representations \cite{zemel13}. These frameworks have been extensively studied and applied to systems with structured data and specific use cases, such as the canonical examples of predicting default in financial lending \cite{kumar22}, predicting recidivism in criminal justice \cite{angwin16}, and coreference resolution in natural language \cite{zhao18b}.

There is an open question of how to think about bias fairness with the advent of generative AI and general-purpose large language models (LLMs). LLMs are increasingly used for a multitude of tasks that span both established areas of concern for bias and fairness—such as evaluating resumes in hiring, where the bias literature goes back decades \cite{bertrand04}—and areas less frequently discussed in the extant fairness literature—such as drafting and editing emails \cite{laban23}, answering general knowledge queries \cite{spatharioti23}, and software development \cite{bird22}.

We approach this topic mindful of both the hotly contested issues already present in the fairness literature \cite[e.g.,][]{corbett-davies17} and the challenges that other ascendant paradigms, such as information access systems \cite{ekstrand22a}, have already presented for the ideal of fairness. For example, it is clear from the extant literature that multiple group fairness metrics, such as those defined by rates of false positives and false negatives \cite{chouldechova17, kleinberg16} or demographic parity and calibration \cite{kleinberg16}, cannot be simultaneously achieved in real-world environments, even to an approximation.

We develop the stronger claim: fairness in the rigorous sense defined by these frameworks for narrow use, even on a single nontrivial metric, is intractable with general-purpose LLMs. The inherent challenges would persist regardless of advances in empirical methods, but we present future directions in light of them. Specifically, we make the following arguments:

\begin{itemize}
    \item Fairness through unawareness of sensitive attributes is made impossible by the unstructured training data and limited transparency of LLMs (\Cref{sec:unaware}).
    \item Standards for the fair treatment of content producers can be rendered obsolete by the LLM capacity for large-scale consumption and redistribution of content (\Cref{sec:producer}).
    \item General-purpose LLMs cannot be made fair across many contexts because of the combinations of populations, use cases, and other factors that impose different fairness requirements (\Cref{sec:context}).
    \item Fairness does not compose, and LLM development and deployment involve the composition of different models, each with their own fairness challenges (\Cref{sec:compose}).
    \item There is much important and tractable work to be done on LLM fairness, particularly in crafting standards of developer responsibility, refining in-depth methods for context-specific evaluation, and building scalable evaluations that iterate through participatory design and using AI capabilities to scale up to the multitude of real-world contexts (\Cref{sec:implications}).
\end{itemize}

\section{Approach}
\label{sec:approach}

In order to assess the compatibility of LLMs with fairness frameworks, we considered each of the fundamental affordances of the LLM paradigm alongside each of the fairness frameworks. We see this as a broadly promising approach to examine new AI affordances across existing sociotechnical frameworks (e.g., agency \cite{sturgeon25}, deterrence theory \cite{hendrycks25}).

First, at the technical level, we observe that LLMs have exceptional flexibility. It is increasingly clear that a wide range of content can be represented in LLM-suitable natural language. LLMs are increasingly multimodal, such as the capability of GPT-4 \cite{openai23d} to receive text, visual, audio, or mixed-modality input. LLMs lack the self-evident use case or even a relatively narrow set of use cases that have grounded prior work within these fairness frameworks. Recent work has demonstrated the need for metrics applicable to real-world deployment contexts and capable of iterative refinement as systems evolve \cite{lum24, wallach25, weidinger25}.

Second, at the social level, our analysis foregrounds the multitude of diverse stakeholders in LLM systems and their continuously evolving relationships. As discussed in \Cref{sec:producer}, there are developers: people and organizations who create datasets, curate datasets, develop models, deploy and manage models, and build downstream user-facing applications; there are users: subjects on which content produced by the system is based; and there are producers of content, such as owners of websites in the context of a search engine. In general, while our critiques are leveled at the applicability of technical frameworks, they echo the many challenges reported by practitioners from real-world deployment \cite{madaio22}.

An illustrative problem that arises with many stakeholders is information asymmetry. Without information from developers (e.g., architecture details, training data), users and third parties have limited ability to conduct thorough evaluations. For a concrete example, consider the February 2024 public controversy in which Google's frontier LLM, Gemini, was found to diversify race and gender appearances in images even when prompts specified historical settings that would be of a particular race and gender, such as soldiers and political figures in American and European historical settings that were almost exclusively men of European descent \cite{milmo24}. While there is much to be debated in how race and gender should be portrayed in image generation, third parties bemoaned the lack of information on the mechanisms by which these images were generated, and the current LLM fairness literature does not fully address such complex cases with diverse stakeholders.

\section{Recent work on LLM fairness}
\label{sec:recentwork}

Interest in LLMs has accelerated in recent years as models such as ChatGPT, Claude, and Gemini have become more pervasive in everyday life, including sensitive contexts such as health and hiring. This has motivated research into many safety and ethical issues. While this paper is not intended as a comprehensive literature review, we first briefly review the recent work in machine learning and NLP research on bias and fairness in LLMs.

\subsection{Association-based fairness metrics}

Two recent reviews of this nascent literature \cite{gallegos23, li24a} enumerate a variety of fairness metrics that each constitute an association between a feature of the embedding space or model output and a sensitive attribute. NLP research in this area includes disparities of sentiment and toxicity in Wikipedia sentence completion across the profession, gender, race, religion, or political ideology of the article subject \cite{dhamala21}, the tendency to generate violent words after a phrase such as ``Two muslims walked into a'' [sic] \cite{abid21}, and variation in the topics introduced when completing sentences from fiction novels \cite{lucy21}. Other approaches include creating datasets of LLM text continuations that include stereotypes, demeans, or otherwise harms in ways related to gender and sexuality \cite{fleisig23}; evaluating an LLM used for the conventional machine learning task of predicting outcomes based on a text-converted tabular dataset \cite{li23}; recommending music or movies to a user who specifies their sensitive attribute, such as race or religion \cite{zhang23a}; and testing whether the model gives the same ``yes'' or ``no'' answer when asked for advice by users who specify their gender \cite{tamkin23}.

However, a lack of disparities in these test cases would not constitute fairness as conceptualized in technical frameworks or in other fields such as philosophy \cite[e.g.,][]{binns21}. For example, within the scope of group fairness, which uses conditional equivalencies of model output across sensitive attributes, the simplest notion—unconditional equivalence—is known as demographic parity. Demographic parity is an important metric to study, but achieving it (i.e., zero disparity) is rarely, if ever, viewed as achieving fairness. While the popular benchmarks that have been applied to LLM-generated text to date, such as WinoBias \cite{zhao18b} and BBQ \cite{parrish22}, capture important information about the associations between generated text and sensitive attributes, strong model performance does not constitute fairness per se. Indeed, even without considering the technical fairness frameworks, the limitations of these benchmarks as proxies for issues such as stereotyping is well-established \cite{blodgett21, lum24}. There is little reason to think that the disparity measures, which are the most common fairness metrics in NLP, serve as sufficient proxies for the fairness frameworks, even with narrow-purpose AI.

Extant work on LLMs has touched on the technical fairness frameworks, but that has typically been in a highly constrained manner. For example, while \citet{li24a} briefly discussed counterfactual fairness, they only did so by summarizing two papers that merely perturb the LLM input, such as by converting Standard American English to African American English \cite{liang23}, which does not acknowledge or address the inherent challenges we present in \Cref{sec:context} of how counterfactual fairness and other metrics fail to generalize across populations and how realistic counterfactuals would not merely vary in writing style or any other features directly observable in the text. Our work, in contrast, critiques the assumption that bias and fairness can be so easily measured.

\subsection{Empirical challenges}

Extant work has articulated significant challenges in achieving LLM fairness, but it has said little about the fairness frameworks that are used to measure and guarantee fairness in conventional machine learning and NLP applications. \citet{gallegos23} and \citet{li24a} overview several issues, such as the need to center marginalized communities \cite{birhane22, blodgett20} and to develop better proxies by bridging the divide between intrinsic and extrinsic bias metrics \cite{goldfarb-tarrant20}. While we cannot presently cover all of the recent work on LLM fairness, including more recent reviews such as \citet{chu24}, we generally note that, even if every empirical challenge were addressed, the inherent challenges that are the focus of the present work would remain. We return to empirical challenges, and means to address them, in \Cref{sec:implications}.

The inherent challenges of LLM fairness have yet to be foregrounded in part because work to date has largely focused on relatively narrow use cases. Often the LLM is applied as a classifier or recommender system in conventional machine learning tasks through the use of in-context learning to produce the conventional output format (e.g., a binary data label) \cite{li23, tamkin23, zhang23a}. It is true that, given the flexibility of LLMs, they could be deployed to any conventional task, but LLMs are not primarily used or advertised as substitutes for conventional, narrow-purpose models. In the following enumeration of inherent challenges, we refer to various studies that provide important conceptual foundations, but our claims are our own synthesis and not extracted directly from prior work.

\section{Inherent challenges of fair LLMs}

\subsection{Unawareness is impossible by design}
\label{sec:unaware}

The framework of fairness through unawareness (FTU), which measures fairness based on whether the model input explicitly contains sensitive attributes, emerged for models built on structured data, typically in which data is organized into variables used for prediction or classification. For example, a financial lending model could use a person’s age, gender, and credit score to make a prediction about loan repayment in which FTU means that ``gender'' is excised from the training data. Legal, policy, and feasibility constraints often lead to the FTU approach in practice. In one of the most widely known allegations of algorithmic discrimination, a group of heterosexual married couples who used the Apple Card noticed after online discussion that each woman was extended a much lower credit limit than her husband. The company managing the Apple Card, Goldman Sachs, defended itself by saying, ``In all cases, we have not and will not make decisions based on factors like gender'' \cite{telford19}.

By design, LLMs are trained on massive amounts of unstructured data, primarily natural language but also visual and audio modalities. FTU is impossible in these contexts because of the pervasiveness of sensitive attributes. Indeed, LLMs are readily able to infer personal characteristics such as the age, location, and gender of an author. For example, \citet{staab24} show that ChatGPT, Claude, and other LLMs can easily guess personal characteristics based on Reddit profiles.

Efforts to remove sensitive attributes can produce incoherence or distortion. For simplicity, we provide an example in which national origin (the sensitive attribute under consideration) is explicitly specified: Consider the sentence, ``Alice grew up in Portugal, so Alice had an easy time on the trip to South America.'' Simply removing Alice’s origin, ``Portugal'' or ``in Portugal,'' would result in an ungrammatical sentence. Other approaches for removing national origin would still result in distortion. Substituting the neutral phrase ``a country'' or ``in a country'' would remove important narrative information, such as the author conveying that Alice visited Brazil, the only South American country in which Portuguese is an official language. The story may go on to describe Alice's global travel, in which her national origin plays an important role in how she reacts to new experiences.

Efforts to remove more implicit sensitive attributes (e.g., of the text author) may result in even more distortion of content, and identifying them may be very challenging and has not been addressed in prior fairness studies (e.g., the aforementioned \citet{liang23}). Consider how relative status can be conveyed through pronoun usage, such as the use of first-person pronouns being more common in groups of lower social status \cite{kacewicz14}. Moreover, in languages with gendered nouns (e.g., Spanish, German), enforcing a notion of gender fairness may require introducing entirely new vocabulary, and if nationality, native language, religion, beliefs, or other attributes of cultural background are considered sensitive, then the corresponding languages, dialects, and subdialects would also be impossible to extirpate. Even with attributes that could be removed without distortion in certain cases, it is infeasible to enforce fairness with respect to all relevant sensitive attributes across a large corpus while retaining sufficient information for model performance. There may also be direct ethical issues with the modification of text, such as authors not consenting to the modifications.

As with the other frameworks, FTU is additionally hindered by the current lack of model transparency. FTU would require that an LLM be documentably unaware of the sensitive information, which would require a level of documentation of training data that is unavailable for any state-of-the-art LLM today—at least to third-party researchers, auditors, and developers. Even with a model such as Llama, for which the weights are shared freely online, there is little public information about training data \cite{dubey24}. Finally, while conventional FTU explicitly leaves out the sensitive attribute, some approaches use the sensitive attribute information to ensure that the model is not even implicitly aware of the sensitive attribute through proxies, such as zip code as a proxy for race and income given the strong predictive relationship \cite{lipton18, pope11}, which would be even more challenging.

\subsection{Producer-side fairness criteria can be rendered obsolete}
\label{sec:producer}

In the literature on fairness in recommender and information retrieval systems, the presence of multiple stakeholders has motivated the multi-sided fairness framework. This framework requires that the system is fair with respect to each group of stakeholders, typically divided into consumers, subjects, and producers of content \cite{abdollahpouri20, burke17, ekstrand22a, sonboli22}. For consumers and subjects (i.e., people or groups who receive the recommendations), there are many possible fairness targets, such as that each consumer or consumer group should receive comparably high-quality recommendations \cite{ekstrand22, ekstrand24, mehrotra17, wang21a}. While there are challenges in measuring quality or utility and what distribution of quality or utility is fair, these are more or less straightforwardly intensified from conventional NLP to LLM use cases.

For subjects, it may be difficult to define, detect, or enforce appropriate fairness metrics, particularly across modalities. For example, there is an open question of whether the target distribution of gender across search engine results for ``CEO'' should be equal representation of men, women, and other genders or a distribution that is weighted towards the gender distribution of CEOs in the consumer’s home location \cite{feng22, karako18, raj22}. These issues are compounded by the lack of clear correspondence between LLM outputs and real-world subjects: Images or texts produced by an LLM-based system often do not correspond directly to particular individuals or even particular levels of sensitive attributes, such as generating images that do not clearly represent a particular race or ethnicity. Note that we consider an image-producing system to still be an LLM given that natural language (e.g., English) is still the primary modality and ``language'' itself can be perceived more broadly to include the encoding and communication of ideas through imagery and other modalities.

There are more complex challenges in multi-sided fairness for producers, also known as providers. The conventional fairness target is typically an equitable distribution of exposure, either in terms of relevance-free metrics that do not consider the relevance of the content to the user—only that there is an equitable distribution—or relevance-based fairness metrics that target an equitable exposure conditional on relevance. This framework can at times transfer directly to LLMs in the context of information retrieval and management tasks. For example, if someone searches for ``coffee shops in San Francisco'' in an LLM chat—as is being incorporated into the ubiquitous modern search engine, Google—producer fairness could be defined in terms of equitable exposure to the different brick-and-mortar coffee shops in San Francisco. Even if the LLM system does not direct users to particular websites, many users will presumably visit the cafes, which provides utility—fairly or unfairly—to the producers.

However, if users search for information via the LLM system, such as asking, ``How are coffee beans roasted?'' then LLMs can entirely circumvent the producers and upend the conventional notion of producer-side fairness. If the LLM system extracts information from websites without directing users to the original source content, then it may be that none of the producers receive any exposure or other benefits in the first place. One way to make sense of this would be to consider the LLM system itself—or the entity that developed, owns, and manages it—as another type of stakeholder, one that takes utility from the producers and renders the conventional producer-side fairness criteria obsolete. This is a particularly important consideration given the ongoing integration of LLMs into search engines, such as OpenAI's SearchGPT \cite{openai24} and Google. While these developers have committed to responsible practices, such as supporting content producers, third-party evaluation can help ensure accountability.

\subsection{General-purpose LLMs cannot be made fair across many contexts}
\label{sec:context}

Much of the excitement surrounding LLMs is based on their general-purpose flexibility across wide ranges of populations, use cases, and sensitive attributes. This flexibility makes many conventional fairness metrics intractable, which we illustrate with the group fairness framework. 

Group fairness metrics, such as demographic parity, equalized odds, and calibration \cite{verma18}, require independence between model classification and sensitive attributes, often conditional on relevant information such as the ground-truth labels that the model aims to predict (e.g., job performance for a model that assists in hiring decisions). In binary classification, these metrics are achieved when equalities hold between ratios in the confusion matrix: equal ratios of predicted outcomes (demographic parity), equal true positive rates and false positive rates (equalized odds), or equal precision (calibration). Recent work includes extensions of these notions, such as prioritizing the worst-off group \cite{diana21}; methods to estimate the sensitive attribute when it is unavailable \cite{kallus21, zhao22, lahoti20}; and methods of enforcement, such as preprocessing \cite{feldman15}.

\subsubsection{Populations and use cases}

LLMs, particularly general-purpose LLMs, present a challenge for group fairness metrics in part because LLMs tend to be deployed across a wide range of input and output distributions. \citet{lechner21} showed that it is impossible for a nontrivial model to perform fairly across all different data distributions, such as regions or demographic groups, to which it might be applied. In current discussions of algorithmic fairness (e.g., recidivism prediction in criminal justice), fairness is typically targeted at a local jurisdiction, which ensures that the model is performing fairly for that location’s particular demographic mix (e.g., age, race) but typically cannot also ensure fairness in different locations. The purpose and use of LLMs makes it infeasible to restrict them to this sort of target population. Interviews with AI practitioners have shown that this complexity is already a substantial challenge in the deployment of conventional AI systems \cite{madaio22}.

In general, it is not clear what an appropriate target population would be on which to detect and achieve group fairness for an LLM. For example, one could bootstrap a predictive model for recidivism prediction from an LLM by instructing it to make a prediction about an individual based on a fixed set of that individual’s characteristics with in-context learning, as \citet{li23} did in predicting the label of a text-converted tabular dataset. However, the data on which that LLM had been trained does not admit an identifiable target population because a corpus of text is not a structured database comprising people and their individual characteristics. An LLM may be trained in part on structured databases, but the output of the model for any such predictions is also based on the wide scope of unstructured training data. This is compounded when the LLM is deployed across many use cases within each population.

Generalization across populations and use cases is also a concern for fairness frameworks other than group fairness because of the wide range of data, use cases, and social contexts at play in LLM use \cite{rauh22}. For two examples: First, individual fairness requires that the model output is Lipschitz continuous with respect to the model input \cite{dwork11}. In this case, it is not clear what similarity metrics could be reasonably applied across the multitude of populations or use cases. If context-specific metrics were applied, it is still left undetermined how these could be judiciously selected and guaranteed.

Second, counterfactual fairness requires that the model would have produced the same output for an individual if they had a different level of the sensitive attribute \cite{kusner17}. However, it is often difficult to identify the causal structure of the data-generating process in even a single task, and it would be an immense challenge for a single model to account for all of the many different contextual factors that determine counterfactuals or other causally distinct outcomes across the varying populations and use cases.

\subsubsection{Sensitive attributes}

Given the issues discussed in \Cref{sec:unaware}, it may not be tractable to exclude sensitive attributes from training data, and each of the populations and use cases can require fairness metrics to be enforced for a different set of sensitive attributes. The effort required increases combinatorially with the importance of intersections of sensitive attributes \cite{himmelreich_intersectionality_2024}.

This is a challenge for the group fairness metrics already defined, but the issue is particularly salient for the popular ideal of fair representations, which requires that data representations do not contain information that can identify the sensitive attributes of individuals \cite{zemel13}.

In the fair representations framework, a system first maps the dataset of individuals being represented to a probability distribution in a novel representation space, such that the system preserves as much information as possible about the individual while removing all information about the individual’s sensitive attribute. The most well-known example of this approach is \citet{bolukbasi16}, which rigorously documented gender bias in Google News word embeddings, namely an association between occupations and a gender vector (e.g., $\vec{\textit{he}} - \vec{\textit{she}}$), such that ``computer programmer'' was coded as highly male while ``homemaker'' was coded as highly female \cite[see][for a review of more recent work]{sesari22}.

Researchers have developed a number of debiasing approaches focused on a particular sensitive attribute dimension, such as zeroing the projection of each word vector onto the dimension \cite{bolukbasi16} or training the model to align the sensitive attribute dimension with a coordinate of the embedding space so that it can be easily removed or ignored \cite{zhao18}. However, \citet{gonen19} showed that such approaches ``are mostly hiding the bias rather than removing it'' because word pairs tend to maintain similarity, reflecting associations with sensitive attributes in what \citet{bolukbasi16} call ``indirect bias.''

Achieving fairness in one LLM context may be contingent on alteration of the statistical relationships between the context-specific sensitive attribute and other features of the data, particularly the removal of information. For example, one may wish to exclude gender information from financial lending decisions, but gender information may be necessary for other tasks, such as drafting or editing an email about a real-world situation that has important gender dynamics that the sender hopes to communicate to the receiver. Moreover, variables closely associated with gender, such as biological sex and pregnancy status, may be essential factors in medical decision-making. In general, attempts to debias for one context may remove or distort important information for another context.

The naive approach of debiasing the model with respect to the union of all potential sensitive attributes—even if it were empirically feasible—would likely be too heavy-handed, leaving the model with little information to be useful for any task. To effectively create a fair LLM for every task, even for only its most important sensitive attributes, one would need to act upon the parameters of the model with surgical precision to alter the relationship between variables only when the model is instantiated for a specific task. This is infeasible with current methods, such as supervised fine-tuning, and currently we do not have robust techniques to debias even a single problematic relationship without incidentally obfuscating it or problematizing other relationships. The game of fairness whack-a-mole appears intractable, dashing hopes of cross-context debiasing.

\subsection{Fairness does not compose, but fairness-directed composition may help}
\label{sec:compose}

Whether a model’s behavior is fair or desirable largely depends on how the model's output will be used. In many modern AI systems, the output of one model is often used as the input to another model, but this process—known as ``composition''—is difficult because fairness does not compose: a fairness guarantee for each of two models is not a fairness guarantee for a system composed of the two models—a point made most explicitly by \citet{dwork19}. Ensuring fairness is particularly challenging when the different systems—such as \mbox{OpenAI's ChatGPT} \cite{openai22} and DALL-E, OpenAI's primary text-to-image model \cite{ramesh21}—operate with different modalities or training data. In the case of Google's Gemini model, the aforementioned February 2024 controversy was compounded by the difficulty of identifying how the text input was related to the image output \cite{milmo24}.

However, it may be possible to use the aforementioned flexibility of general-purpose LLMs to create fair context-specific model compositions, enforcing fairness ideals in seemingly intractable contexts. This is due to, first, the LLMs’ ability to account for many patterns in data not immediately observable by human model designers—which is much of the reason for excitement about LLMs in recent years—and, second, the instruction tuning that allows them to obey natural language input. Eventually, they may be able to obey a general command to enforce context-specific fairness. Many advances in LLM capabilities can be conceptualized as encouraging the model to improve its own output. For example, chain-of-thought prompting \cite{wei22} encourages the model to first produce text that takes an incremental reasoning step towards its target, which can increase performance by allowing the later token generations to build on the logical reasoning text that the model has already generated, which has then become part of its input.

One can view many approaches to instruction tuning as a composition of an ethics-driven model with the primary LLM. The most popular approaches to alignment and safety, currently Reinforcement Learning from Human Feedback \citep[RLHF;][]{ouyang22} and Direct Preference Optimization \citep[DPO;][]{rafailov23}, compel the model towards human-provided preference data, and some other approaches, such as constitutional AI \cite{bai22} and \texttt{SELF-ALIGN} \cite{sun23}, steer the model towards LLM-generated proxies of human preferences.

While AI-assisted fairness is an interesting possibility, it could easily make the situation worse if attempted before models have the capability to do this safely. The fairness-enforcing model could double down on its own blindspots, particularly those that are not yet sufficiently well-understood or appreciated by the human developers such that they can be guarded against. Recent approaches focus on model ``self-correction.'' There is skepticism that models can currently do this well, but \citet{ganguli23} show impressive results on bias and discrimination benchmarks ``simply by instructing models to avoid harmful outputs.''

\section{Implications and future research}
\label{sec:implications}

We conclude with a brief discussion of how to move forward with building fair AI systems and researching LLM fairness in light of these challenges.

\subsection{Developer responsibility}

Fairness issues manifest throughout the end-to-end pipelines from AI model design and training to model deployment and long-term effects. Users, regulators, researchers, and auditors have historically been well-positioned to evaluate the later stages of this pipeline, but there are substantial challenges for their efforts to understand the earlier stages, hindering efforts towards goals such as fairness through unawareness is infeasible (\Cref{sec:unaware}). LLM developers have a responsibility to support users and third parties. For researchers and other third parties to conduct grounded evaluations, companies that deploy LLMs should share information on actual usage and how the systems respond to real prompts from real users \cite{caliskan24, lum24}. The challenges of unstructured data, producer equity, and diverse contexts suggest a need for LLM developers to work closely with third-party researchers, policymakers, end users, and other affected stakeholders in a participatory and context-informed design process \cite{muller93}.

Modern generative AI systems are trained with unprecedented amounts of natural language and multimodal data. In addition to lacking transparency of training data, the extensive data scraping efforts raise concerns about copyright and intellectual property law \cite{abbott23, chu24a}. If a user or advertiser pays a search engine, that could be unfairly extracting value from both producers (as discussed in \Cref{sec:producer}) of the search result content as well as producers of training data for the underlying LLM. The tendency of general-purpose LLMs to intake extremely large training datasets also raises concerns about the filters used for selection, such as ``quality'' text filters that may disproportionately exclude certain voices \cite{lucy24}. Transparency challenges are compounded by the lack of evaluation infrastructure for LLMs, unlike transportation, aerospace, pharmaceuticals and other fields with mature evaluation regimes established through decades of institutional investment \cite{weidinger25}.

\subsection{Context-specific evaluations}

Building better general-purpose AI systems and measuring their fairness—even if we cannot say that the system is generally fair—will require articulating specific connections to real use cases and corresponding harms, adapting technical frameworks to the specificity of a particular context \cite[e.g.,][]{anthis23a, blandin24}. With the challenges of translating and composing fairness across models and contexts (\Cref{sec:context}), it is unlikely that any ``trick tests,'' such as coreference resolution of gendered pronouns, will provide satisfactory evidence for or against LLM fairness \cite{lum24}. There has been a dearth of proper contextualization in the fairness literature \cite[e.g.,][]{alertubella23, blodgett20}, and the intractability of generalized fairness adds weight to this critique.

Bias is often present from pretraining data, such as large-scale internet corpora, but the interactive feedback loops of LLM prompt engineering, custom instructions, and supervised fine-tuning risk amplifying biases by further shifting the context in which the LLM operates. Users who speak low-resource languages already face lower model performance \cite[e.g.,][]{openai24a}, a challenge that can be compounded by a limited ability to iterate on prompting strategies. As a user continues to interact with a system, even small biases can be amplified, and guardrails can erode—though empirical research is needed on how this manifests with modern LLM use. As LLMs create more content and synthetic data that is used in training new systems, they can exacerbate the deterioration of public goods, including ``enshittification'' as people become locked into online platforms and content quality deteriorates \cite{doctorow25}.

In 2024, the risks of LLM extensibility and customization became salient in public policy debates, such as the European Union AI Act and California's SB 1047, the \textit{Safe and Secure Innovation for Frontier Artificial Intelligence Models Act}, which would ostensibly require that LLM developers implement safety guarantees that include the prevention of misuse and the capability to promptly shut down the system if necessary. Critics have argued that LLM developers cannot make such guarantees because LLMs are inevitably deployed in new and unexpected contexts and are able to be substantially modified by users and third parties, particularly with the development of open source models that allow academics and independent developers to research and innovate.

\subsection{Scalable evaluation}

Today, developing fairness metrics for a single context requires substantial effort to iteratively study harms and develop mitigations. The difficulty increases combinatorially with the variety of populations, use cases, and sensitive attributes and their intersections \cite{himmelreich_intersectionality_2024}, across which any realistic amount of effort is insufficient. Intensive strategies that interview and co-design with stakeholders can supply clarity \cite{madaio_assessing_2022}, but ideally there would be more scalable evaluations that—while LLM fairness guarantees are intractable—can meaningfully support fairness across the many different LLM use cases.

We believe there is an exciting and largely untapped opportunity at the intersection of technical fairness frameworks (and related technical frameworks, such as privacy) and scalable human-AI alignment. For example, expanding on our argument in \Cref{sec:compose}, the ethics-driven methodologies of RLHF \cite{ouyang22}, DPO \cite{rafailov23}, constitutional AI \cite{bai22} and \texttt{SELF-ALIGN} \cite{sun23} can each incorporate technical frameworks and context-adaptive methods. This can include feeding contextual information to a model that meaningfully synthesizes it in a human-like way and adjusts accordingly. There can also be AI-automated evaluation pipelines, including ``LLM-as-a-judge'' \cite{kanepajs_what_2025, zheng_judging_2023} fairness rubrics, high-quality simulations of human data \cite{anthis_llm_2025}, and the generation and validation of tests or simulated user queries \cite{sturgeon25} that probe fairness at scale.

One of the primary challenges of this research direction will be accounting for bias from each of the inputs, such as the reward model in RLHF that inevitably comes from particular people, raising questions about to whose values LLMs are being aligned \cite{gabriel_artificial_2020}; which of their values LLMs are being aligned, such as whether ratings are based on ``helpfulness,'' ``honesty,'' or ``harmlessness'' \cite{askell21, liu24}; and at what times that input is elicited, given changes in values over time \cite{carroll24}. One could also utilize interpretability tools \cite[e.g.,][]{singh23, nanda22} that allow participatory and iterative exploration of how bias manifests. Again, bias can manifest through these tools, such as the quality of interpretation provided to different users in different contexts or variation in the tendency of LLMs to refuse user requests across groups \cite{wester24}. On the other hand, if interpretability tools succeed in providing relevant information, this can be used to make fairer models as they allow for more context-specific adjustments. The potential for harm and benefit will depend in part on the quality of future interpretability tools, especially because RLHF and related techniques tend to incorporate myopic and biased input \cite{casper23}, which could lead to overconfident or otherwise inaccurate explanations.

It remains true that humans cannot feasibly scale manual efforts to the scope of LLMs despite the risk of ``bias all the way down'' when using AI tools to address AI issues. Moreover, AI tools have the unique advantage that their capabilities will scale alongside the risks from powerful AI systems. AI tools should be considered across not just fairness but the host of AI issues (e.g., privacy, mental health) to make progress towards safe and beneficial general-purpose AI.

\section{Conclusion}

Work to date has measured associations and disparities as LLMs provide different output when different demographics are specified explicitly (e.g., ``White'' and ``Black'') or implicitly (e.g., dialects). However, these cannot substitute for more rigorously developed fairness frameworks, such as group fairness and causal fairness. When we consider these frameworks, the inherent challenges render general fairness impossible for an LLM. While these limits are underappreciated in the current literature, there are promising research and practical directions in standards of developer responsibility, context-specific evaluations, and scaling evaluation in ways that cautiously utilize general-purpose AI while mitigating the amplification of moral issues within those systems as well.

\section*{Limitations}

Given the complexity and opacity of today's deep neural networks, it is difficult to formally analyze their capabilities and limitations. The preceding claims and analysis were not developed mathematically, and the outcome of such analysis would depend on particular assumptions and operationalizations. It is also possible that compelling new technical frameworks, perhaps developed specifically for general-purpose LLMs or other general-purpose systems, will circumvent the inherent challenges we described. Finally, while we believe it is important to lay out a conceptual foundation of what is and is not possible, there are many open empirical challenges that we have not addressed in this work, particularly the quantification of how much fairness metrics can be partially satisfied in real-world settings and the development of scalable methods for context-specific alignment with fairness and other social values.

\section*{Acknowledgments}

We are particularly grateful to Alexander D'Amour for his significant contributions to this paper. We also thank Micah Carroll, Eve Fleisig, members of the Knowledge Lab at the University of Chicago, and members of Stanford NLP Group for helpful feedback and suggestions.

\bibliography{references}

\end{document}